\documentclass[conference]{IEEEtran}
\IEEEoverridecommandlockouts

\usepackage{cite}
\usepackage{amsmath,amssymb,amsfonts}
\usepackage{algorithmic}
\usepackage{graphicx}
\usepackage{textcomp}
\usepackage{xcolor}
\usepackage{booktabs}
\usepackage{multirow}

\usepackage{caption}
\begin{document}

\title{Feeling Terrain Before Crossing: World Models for Off-Road Navigation\\
    \thanks{All authors are with the Department of Electrical \& Computer Engineering, Seoul National University, Republic of Korea.
    {\tt\footnotesize \{pingpang, sgr01086, hoons21, calvin1225, trr0631, mychoco333, sseo\}@snu.ac.kr}}
}


\author{
\IEEEauthorblockN{E-In Son, Dong-Wook Kim, Ji-Hoon Hwang, Kangsun Lee, Jisung Bae, Jung-Taak Kim and Seung-Woo Seo}

}

\maketitle

\begin{abstract}
Navigation world models plan by foresight, predicting the future that each
candidate action sequence produces and selecting the best, rather than mapping
observations to actions directly. Unlike urban settings where a predicted scene
is a sufficient proxy, off-road navigation hinges on the robot--terrain
interaction, so the prediction must cover not only what the camera will see but
what the robot will feel. However, existing scene-focused models do not predict
how much the robot will slip, tilt or shake along a planned trajectory.
Proprioception captures these dynamics directly and, when used as input,
improves the prediction of the physical future. We present Feel-WM, the first
off-road navigation world model that conditions on proprioception and predicts
what the robot will feel alongside what the camera will see. The physical future
takes the form of a future proprioceptive state and a failure risk, both learned
from the robot's own experience without human labels. The planner rolls out the
physical future alongside the scene and weighs the predicted failure risk
against goal similarity in a separable score. Experiments on real off-road data
and in simulation demonstrate that Feel-WM outperforms visual-only navigation
world models in open-loop planning and closed-loop rough-terrain navigation
across wheeled and legged platforms. Deployed on a Husky on mountain trails,
Feel-WM plans onboard, predicts rough ground ahead and steers around it,
completing courses that an end-to-end policy fails.
\end{abstract}

\section{Introduction}
\label{sec:introduction}

Vision-based navigation has largely relied on end-to-end policies trained on
demonstrations to map camera observations to
actions~\cite{shah2023vint, sridhar2024nomad, kim2026canonnav}.
Navigation world models introduce an intermediate prediction step. Rather than
selecting an action directly from the current view, they predict the future of
each candidate action sequence and select the sequence whose predicted future is
closest to the goal~\cite{bar2025navigation, zhang2025efficient, zhang2026rae, zhou2025dino-wm}.
Predicting before acting is worth even more on off-road terrain, where an
inappropriate action can cause the robot to overturn or become stuck. A robot
that crosses a slope safely at low speed can roll over on the same slope when it
enters faster with its body already tilted~\cite{RothP-RSS-25, zhao2026learning}.

\begin{figure}[t]
  \centering
  \includegraphics[width=\linewidth]{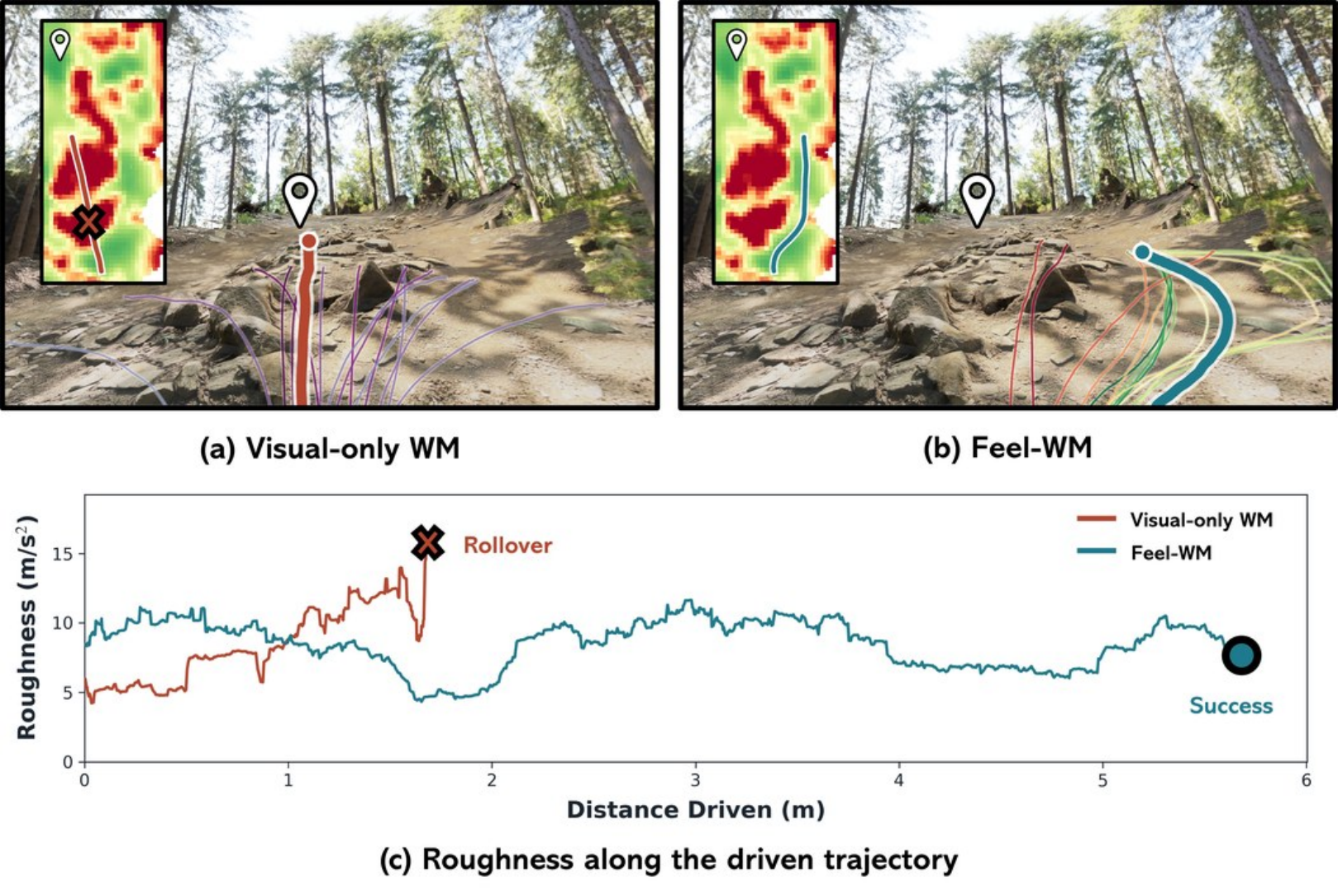}
  \caption{\textbf{Same start, goal and terrain, two futures.} (a)~The visual-only world
  model drives across the rock field; (b)~Feel-WM goes around. Thin lines denote the
  candidates and the thick line the driven trajectory, which the insets place on a roughness map;
  $\times$ marks the rollover. (c)~Roughness along each driven trajectory.}
  \label{fig:teaser}
\end{figure}

Yet the future that a navigation world model predicts is what the camera will see.
In urban and indoor settings, this view suffices to judge where a plan ends. On
off-road terrain, whether a plan succeeds depends on the robot--terrain
interaction along the way, and a predicted view captures it only indirectly. The
interaction appears in the robot's own motion, which the model neither observes
nor predicts. Recent navigation world models widen what they predict, from how
nearby people will move~\cite{hu2026navthinker} to a terrain map for off-road
driving~\cite{yang2026terrainformer}, but the robot stays outside.

Proprioception brings the robot into the prediction. An IMU and wheel or joint
encoders measure the robot's own motion, its state and how much it slipped,
tilted and vibrated. Several lines of off-road work exploit these signals:
forward dynamics models predict the response of the vehicle from its
state~\cite{RothP-RSS-25}, self-supervised costmaps learn terrain cost from what
the vehicle felt~\cite{castro2023does}, and legged traversability estimators
infer terrain properties through foot--terrain
interaction~\cite{frey23fast, elnoor2024pronav}. These works motivate an
off-road navigation world model that predicts what the robot will feel alongside
what the camera will see.

\begin{figure*}[t]
  \centering
  \includegraphics[width=0.85\linewidth]{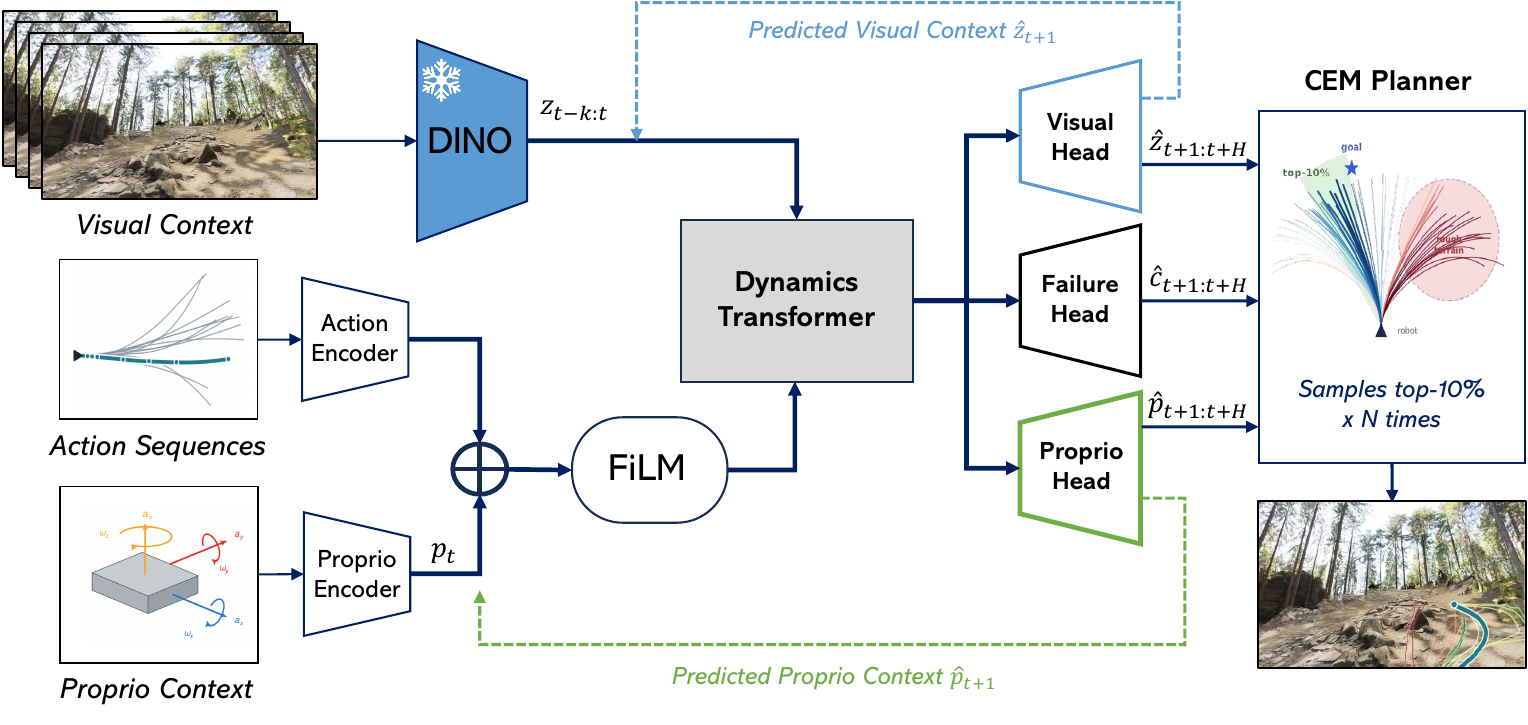}
  \caption{\textbf{Overview of Feel-WM.} FiLM injects the action and the current
  proprioceptive state into the frozen DINOv2 patch tokens, and the dynamics transformer
  predicts the next visual tokens, a failure risk and the next proprioceptive state. Dashed:
  the predictions that re-enter at the next step.}
  \label{fig:method}
\end{figure*}

We present Feel-WM, the first off-road navigation world model to take
proprioception as input and predict the physical future---what the robot will
feel before crossing the ground ahead. A proprioceptive encoder conditions the
dynamics model on the robot's current state. Two heads predict the robot's
future proprioceptive state and a failure risk. Since the prediction depends on
the robot state and the candidate actions, the model distinguishes a safe
crossing from a dangerous one on the same ground. On recorded off-road driving,
the proprioceptive input improves this prediction over the horizons the planner
uses. At planning time, the planner rolls out both predictions for each
candidate and weighs the predicted failure risk against goal similarity in a
score with two separable terms (Fig.~\ref{fig:teaser}).

Feel-WM outperforms visual-only navigation world models in open-loop planning
on recorded off-road driving and in closed-loop navigation with simulated
wheeled and legged robots on rough terrain. Deployed on a Clearpath Husky,
Feel-WM plans on an onboard Jetson Orin and completes off-road courses that an
end-to-end policy fails.

\section{Related Work}
\label{sec:related}

\subsection{Navigation World Models}
Visual navigation policies are trained end-to-end on demonstrations to map
camera observations to actions. ViNT~\cite{shah2023vint} trained a Transformer
policy on diverse multi-robot data for image-goal navigation, and
NoMaD~\cite{sridhar2024nomad} extended it with a diffusion decoder that
represents multiple plausible actions in a scene. Navigation world models
instead predict the future of each candidate action sequence and select the
best. NWM~\cite{bar2025navigation} introduced this paradigm by generating future
egocentric views with a conditional diffusion transformer and ranking candidates
by the similarity between predicted views and the goal image. Subsequent work
has pursued two directions: reducing rollout costs and expanding prediction
beyond future visual observations.

\textbf{From pixels to features.} AstraNav-World~\cite{chen2025astranav} stayed
in pixel space, coupling a video diffusion model to a policy. Generating frames
is expensive, however, and later models have traded fidelity for speed. A
one-step world model~\cite{shen2026efficient} predicts the next observation in a
single pass, and DINO-WM~\cite{zhou2025dino-wm} ran the rollout over frozen
DINOv2 features and planned toward goal features without task-specific
training. ReL-NWM~\cite{zhang2025efficient} and RAE-NWM~\cite{zhang2026rae}
followed with a compact latent and a dense representation. Although these
models differ in how they represent the future, what they predict remains the
scene. DINO-WM concatenates the agent's proprioceptive state to the latent when
one is available, but as the agent's configuration in tabletop and maze tasks,
not as a measurement of how the terrain acts on the robot.

\textbf{Task-aligned outputs.} Each model extends prediction to suit its target
setting. NavThinker~\cite{hu2026navthinker} predicts nearby pedestrians'
trajectories for social navigation, and WMNav~\cite{nie2025wmnav} uses a
vision-language model to guide rollouts toward an object goal. World-action
models~\cite{azuma2026navwam, yang2026wam} incorporate actions into the
prediction and eliminate the need for a separate planner.
TerrainFormer~\cite{yang2026terrainformer} predicts a bird's-eye-view map of
traversability and elevation for off-road driving. These works extend
navigation world models beyond future visual observations, but none conditions
on the robot's proprioceptive state or predicts its physical future.

\subsection{Proprioception in Off-Road Navigation}
On rough terrain, the effect of an action depends on the robot's current state
as much as on the terrain ahead. Roth et al.~\cite{RothP-RSS-25} condition a
forward dynamics model on proprioceptive history and surrounding geometry to
predict a legged robot's future position and failure probability. A
sampling-based planner then ranks candidate trajectories by the predicted
failure probability rather than a hand-tuned cost. Lee et
al.~\cite{lee2024learning} carry the same conditioning to wheeled platforms,
predicting position, orientation and bumpiness from terrain and vehicle state.
Proprioception also provides supervision: Castro et al.~\cite{castro2023does}
learn an off-road costmap from recorded vehicle vibrations, and
ProNav~\cite{elnoor2024pronav} estimates the traversability of the terrain under
a legged robot from joint and IMU signals alone. The forward dynamics models
predict the robot's physical future from a geometric terrain representation
toward a position goal. Feel-WM predicts it from camera features toward an image
goal, alongside the future view.

\subsection{Traversability Estimation and Risk-Aware Planning}
In off-road navigation, traversability estimation helps planners identify
failures by assigning terrain-dependent costs. Frey et al.~\cite{frey23fast}
estimate traversability online from images, supervised by the robot's own driving, and
Velociraptor~\cite{triest2024velociraptor} produces cost and speed maps without
human labels. TAO~\cite{yoo2024traversability} jointly optimizes a path and a
speed profile over traversability maps in mountainous terrain. More recently,
MAT~\cite{zhao2026learning} models traversability as a function of the crossing
velocity. For example, a ditch stops a slow vehicle but can be crossed with an
accelerated jump. We build on the same premise: the same ground can be passable
in one robot state and impassable in another. However, this velocity-conditioned
map does not account for the robot's broader proprioceptive state or how it
evolves during a traversal.

These lines of research address two complementary requirements for
rough-terrain planning: predicting the future scene and the robot's physical
future. Feel-WM brings them together. To the best of our knowledge, it is the
first world model for off-road navigation that conditions on proprioception,
predicts the physical future together with the scene, and plans over both.

\section{Methods}
\label{sec:method}

\subsection{Problem Setup and Overview}
\label{sec:method:formulation}

We consider goal-conditioned navigation on rough terrain. Given a goal image,
the robot must select actions that reach the goal without failing along the
way.

A navigation world model plans by predicting future
observations~\cite{bar2025navigation, zhang2025efficient}. A frozen encoder
maps each camera frame to a feature map $z_t$, and a world model $W$ predicts
future features conditioned on a candidate action sequence:
\begin{equation}
  \hat{z}_{t+1:t+H} = W\!\left(z_{t-k:t},\; a_{t:t+H-1}\right),
  \label{eq:nwm}
\end{equation}
where $a_t = (\Delta x, \Delta y, \Delta\psi, \Delta t)$ specifies a body-frame
pose change over a duration $\Delta t$, measured from the recorded poses in
training; $k{+}1$ is the number of context frames and $H$ is the rollout
horizon. The planner ranks candidate sequences by the
similarity between the predicted future features and the goal image features.

On rough terrain, what an action does to the robot depends on the robot's
current state as much as on the terrain ahead~\cite{RothP-RSS-25}. However,
Eq.~\eqref{eq:nwm} neither conditions on the robot's proprioceptive state nor
predicts its evolution. Feel-WM extends both the inputs and the outputs:
\begin{equation}
  \left(\hat{z},\, \hat{p},\, \hat{c}\right)_{t+1:t+H}
  = W\!\left(z_{t-k:t},\; p_{t-k:t},\; a_{t:t+H-1}\right),
  \label{eq:ours}
\end{equation}
where $p_t$ summarizes the robot's proprioceptive measurements over a window
ending at frame $t$. Since the model reads the recent proprioceptive history,
it predicts different futures for the same ground depending on the robot's
state, for example a slope entered slowly or at speed. Alongside future visual
features, the model predicts proprioceptive features $\hat{p}$ that describe
body vibration, rotation and wheel slip. The model also predicts a failure risk
$\hat{c}$, defined as the probability of failure during each rollout step.
Fig.~\ref{fig:method} provides an overview.

\subsection{Feel-WM}
\label{sec:method:architecture}

Feel-WM operates entirely in feature space. A frozen DINOv2
encoder~\cite{oquab2024dinov} maps each frame to patch tokens, which serve as
the visual representation for prediction and planning without image decoding.

\paragraph{Reading the robot's state}
Proprioceptive measurements arrive at a much higher rate than camera frames,
and their temporal evolution captures motion patterns that a single sample
cannot. A small 1-D convolutional network processes the high-rate measurement
window ending at frame $t$ (inertial accelerations, angular rates, body
velocities, attitude and actuator readings) and pools it into a single
embedding $p_t$. The step's action and $p_t$ condition the current frame's
patch tokens through feature-wise linear modulation (FiLM~\cite{perez2018film}).

\paragraph{Predicting from a shared state}
A dynamics transformer with temporally causal attention over the context tokens
produces a shared representation $h_t$ at each step, which feeds three
prediction heads. The visual head regresses the next frame's features for each
patch, as in a visual-only model. The proprioceptive and failure-risk heads
first pool $h_t$ across patches, because what the robot will undergo is a
property of the whole view rather than of one patch. The failure risk therefore
describes a crossing conditioned on the robot's state and action, rather than a
fixed property of a location.

\paragraph{Rolling forward}
An $H$-step rollout applies this process autoregressively. Neither future
observation is available when the rollout is planned, so both streams are
advanced from the model's own predictions: predicted patch features are appended
to the visual context, and the predicted proprioceptive state is projected back
into the conditioning input. After the first step, the model therefore uses its
own predictions to advance the rollout.

\subsection{Training}
\label{sec:method:training}

Feel-WM is trained on the robot's own driving. Proprioceptive targets and
failure labels are computed from the recordings themselves, without human
annotation.

\paragraph{Defining the physical future}
The proprioceptive targets are predefined statistics computed over a short
window centered on the target time:
\begin{alignat}{2}
  &f_{\mathrm{rough}} = \sigma(a_z),
    &\quad &f_{\mathrm{ang}} = \sqrt{\sigma^{2}(\omega_{\phi}) + \sigma^{2}(\omega_{\theta})},
  \nonumber \\[6pt]
  &f_{\mathrm{slip}} = \frac{\bar{v}_{\mathrm{w}} - \bar{v}_{\mathrm{g}}}
                            {\max(\bar{v}_{\mathrm{w}}, \bar{v}_{\mathrm{g}}, \epsilon)},
    &\quad &f_{\mathrm{slope}} = \overline{v_{\mathrm{g}} \sin\theta},
  \label{eq:channels}
\end{alignat}
where $a_z$ is vertical acceleration, $\omega_{\phi}$ and $\omega_{\theta}$ are
roll and pitch rates, $v_{\mathrm{w}}$ is the encoder-derived wheel speed,
$v_{\mathrm{g}}$ is the ground speed from the state estimator and $\theta$ is
pitch; $\sigma$ and $\sigma^{2}$ are the standard deviation and variance over
the window and overlines denote window means. Roughness and angular
instability quantify vertical vibration and variability in roll and pitch
rates, respectively. The slip feature measures the relative discrepancy between
wheel and ground speeds, and the slope feature serves as a proxy for vertical
velocity. All four channels are standardized using z-scores to reduce
differences in scale during regression. Binary failure labels are derived from
the same recordings: in simulation, a step is labeled unsafe within a predefined
interval before a collision, rollover or stall; on TartanDrive~2, when the
vehicle's recorded hazard signal exceeds a fixed threshold.

\paragraph{Objective}
All three losses are optimized jointly from the start of training. The visual loss combines squared error and cosine distance between
predicted and target patch features. The proprioceptive loss applies weighted
regression to $\hat{p}$, and the failure-risk loss applies class-weighted
cross-entropy to $\hat{c}$ to account for the rarity of unsafe steps. The total
objective combines these losses with fixed weights:
\begin{equation}
  \mathcal{L} = \mathcal{L}_{\mathrm{vis}} + \lambda_{p}\,\mathcal{L}_{\mathrm{prop}}
  + \lambda_{c}\,\mathcal{L}_{\mathrm{fail}}.
  \label{eq:loss}
\end{equation}

\paragraph{Training the rollout used by the planner}
We supervise multi-step rollouts using scheduled sampling. To reflect
planning-time operation, where the model recursively consumes its own
predictions, we make a single teacher-forcing decision per training sample
rather than per step. Each sample therefore uses either ground-truth inputs or
its own predictions throughout the rollout. The proportion of autoregressive
samples increases as the teacher-forcing probability decays during training.

We make two additional choices to account for the temporal properties of the
signals. Because the physical channels decorrelate within seconds, a shared
mask restricts both physical losses to steps within a supervision horizon
$H_p$, while the visual loss applies at every step. Each sample also uses a
single sampled step duration $\Delta t$, with action increments rescaled to
preserve their rates. An $H$-step rollout therefore spans a multi-second
lookahead in few recursive predictions, which limits the accumulation of error.

\begin{figure}[t]
  \centering
  \includegraphics[width=\linewidth]{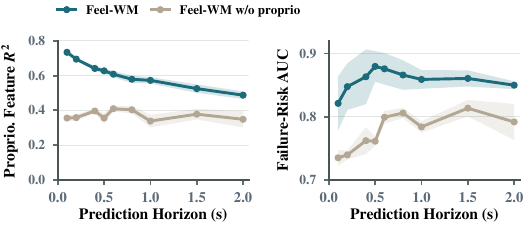}
  \caption{\textbf{Predicting the physical future with and without proprioceptive input.}
  TartanDrive~2 validation, three seeds, mean $\pm$ std. Left: $R^2$ of the predicted
  proprioceptive features. Right: failure-risk AUC.}
  \label{fig:physical}
\end{figure}

\subsection{Planning}
\label{sec:method:planning}

The planner searches for an action sequence using the cross-entropy
method~\cite{deboer2005tutorial}. At each iteration, it samples candidate
sequences, rolls them out through $W$, scores the predicted futures and refits
the sampling distribution to the highest-scoring fraction of candidates. The
search optimizes the pose changes in $a$; each step's duration $\Delta t$ stays
fixed.

The score balances goal similarity against predicted failure risk:
\begin{equation}
  S(\tau) \;=\; \tilde{s}_{\mathrm{goal}}
  \;-\; \lambda_{\mathrm{fail}} \, \tilde{r}_{\mathrm{fail}},
  \label{eq:score}
\end{equation}
where each tilde denotes standardization across the candidate pool by
subtracting the mean and dividing by the standard deviation.
$\lambda_{\mathrm{fail}}$ is therefore a dimensionless trade-off parameter. The
goal term $s_{\mathrm{goal}}$ is the mean per-patch cosine similarity between
the predicted features at the rollout endpoint and the goal image features,
matching the objective used by a visual-only world model. The risk term
$r_{\mathrm{fail}}$ is the mean predicted failure risk $\hat{c}$ over the
rollout. Predicted proprioceptive features condition subsequent rollout steps
rather than entering the score directly.

We retain separate goal and risk terms because the recordings show what the
robot underwent on a given line but not how much risk a mission should accept
for progress. The parameter $\lambda_{\mathrm{fail}}$ controls this trade-off
and can be adjusted without retraining $W$.

\begin{table}[t]
\centering
\small
\setlength{\tabcolsep}{5pt}
\begin{tabular}{@{}lcccc@{}}
\toprule
\multirow{2}{*}{Method} & \multicolumn{2}{c}{TD1} & \multicolumn{2}{c}{TD2} \\
\cmidrule(lr){2-3}\cmidrule(lr){4-5}
 & ADE $\downarrow$ & RTE $\downarrow$ & ADE $\downarrow$ & RTE $\downarrow$ \\
\midrule
NoMaD~\cite{sridhar2024nomad}      & \underline{2.226} & \underline{0.582} & \textbf{3.124} & \textbf{0.755} \\
NWM CDiT-XL~\cite{bar2025navigation} & 5.356 & 1.245 & 5.886 & 1.347 \\
RAE-NWM B/2~\cite{zhang2026rae}  & 5.526 & 1.294 & 5.877 & 1.391 \\
Feel-WM                         & \textbf{1.666} & \textbf{0.551} & \underline{3.439} & \underline{1.016} \\
\bottomrule
\end{tabular}
\caption{\textbf{Open-loop planning on TartanDrive.} 100 paired planning queries per
validation split, 3.2\,s horizon; ADE and RTE in meters. Bold: best per column;
underline: second best.}
\label{tab:openloop}
\end{table}

\subsection{Implementation Details}
\label{sec:method:implementation}

We use a frozen DINOv2 ViT-B/14~\cite{oquab2024dinov} encoder that produces 256 patch tokens per
frame, each with 768 dimensions. The proprioceptive encoder is a 1-D
convolutional network that processes 50 samples from a causal
$0.5\,\mathrm{s}$ window ending at each frame and outputs a 128-dimensional
embedding. The dynamics transformer has six layers, eight attention heads, an
MLP ratio of four and a dropout rate of $0.1$. We set $k = 3$, corresponding to
four context frames, a rollout horizon of $H = 8$ steps and a supervision
horizon of $H_p = 1\,\mathrm{s}$ for the physical losses. Each training sample
uses a single step duration drawn from $\{0.1, 0.2, 0.5, 1.0\}\,\mathrm{s}$.
Each proprioceptive target channel is computed over a $0.5\,\mathrm{s}$ window
centered on the target time.

\begin{figure*}[t]
  \centering
  \includegraphics[width=\linewidth]{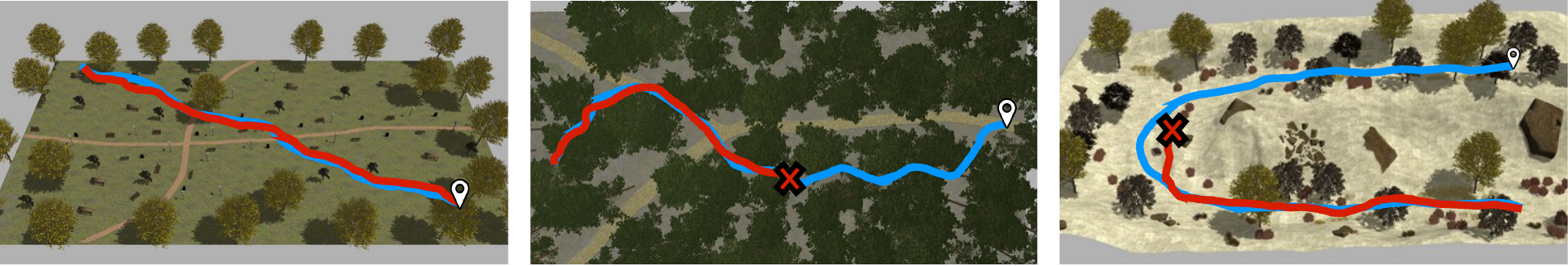}
    \caption{\textbf{Qualitative comparison in Gazebo.} Pins and $\times$ denote the goal and rollover. Trajectories are color-coded as visual-only model (red) and Feel-WM (blue).}
  \label{fig:gazebo}
\end{figure*}

\begin{table*}[t]
\centering
\small
\setlength{\tabcolsep}{6pt}
\begin{tabular}{@{}lllcccc@{}}
\toprule
\multirow{2}{*}{Environment} & \multirow{2}{*}{Route} & \multirow{2}{*}{Method}
   & \multicolumn{2}{c}{\textit{Task performance}} & \multicolumn{2}{c}{\textit{Ride quality}} \\
\cmidrule(lr){4-5}\cmidrule(lr){6-7}
 & & & SR (\%) $\uparrow$ & Prog.\ (\%) $\uparrow$ & Rough.\ (m/s$^2$) $\downarrow$ & Yaw vib.\ (rad/s) $\downarrow$ \\
\midrule
\multirow{9}{*}{Gazebo}
 & \multirow{3}{*}{Park} & Visual-only WM & \textbf{100.0} & \textbf{99.5} & 3.83 & 0.50 \\
 &                                      & Feel-WM w/o prop. & 98.0 & 99.3 & 3.74 & 0.49 \\
 &                                      & Feel-WM & \textbf{100.0} & 99.4 & \textbf{3.71} & \textbf{0.48} \\
\cmidrule(l){2-7}
 & \multirow{3}{*}{Forest}              & Visual-only WM & 54.0 & 80.3 & 4.08 & \textbf{0.52} \\
 &                                      & Feel-WM w/o prop. & 74.0 & 90.3 & 4.07 & 0.74 \\
 &                                      & Feel-WM & \textbf{84.0} & \textbf{95.0} & \textbf{4.00} & \textbf{0.52} \\
\cmidrule(l){2-7}
 & \multirow{3}{*}{Hill}                & Visual-only WM & 58.0 & 74.5 & 3.75 & 0.52 \\
 &                                      & Feel-WM w/o prop. & 68.0 & 82.9 & 3.62 & 0.45 \\
 &                                      & Feel-WM & \textbf{84.0} & \textbf{88.6} & \textbf{3.53} & \textbf{0.41} \\
\midrule
\multirow{6}{*}{Isaac Sim}
 & \multirow{3}{*}{Smooth} & Visual-only WM & 89.7 & 90.7 & \textbf{5.54} & 0.45 \\
 &                                      & Feel-WM w/o prop. & 94.6 & 90.6 & 5.68 & 0.45 \\
 &                                      & Feel-WM & \textbf{98.9} & \textbf{91.3} & 5.62 & \textbf{0.40} \\
\cmidrule(l){2-7}
 & \multirow{3}{*}{Rough}               & Visual-only WM & 67.7 & 72.4 & \textbf{6.46} & 0.73 \\
 &                                      & Feel-WM w/o prop. & 62.0 & 68.9 & 6.53 & 0.77 \\
 &                                      & Feel-WM & \textbf{78.3} & \textbf{77.4} & 6.51 & \textbf{0.65} \\
\bottomrule
\end{tabular}
\caption{\textbf{Closed-loop navigation on rough terrain in two simulators.} 50 episodes
per model per route; 3 Gazebo routes and 13 Isaac Sim environments split by route slope into
7 smooth and 6 rough. Bold: best per block.}
\label{tab:main}
\end{table*}

We set $\lambda_{p} = 0.5$ and $\lambda_{c} = 1.0$, with a weight of $0.1$ for
the cosine term in the visual loss. The teacher-forcing probability remains at
$1$ for the first quarter of training, decays to $0.5$ over the next half and
remains at $0.5$ for the final quarter. We train with AdamW for 20 epochs using
a batch size of 64, a learning rate of $6\times10^{-4}$ and weight decay of
$0.05$. The learning-rate schedule uses one warm-up epoch followed by cosine
decay. The planner refits its sampling distribution to the top $10\,\%$ of
candidates in each pool and uses the mean of the final elite set for execution.

\section{Experiments}
\label{sec:experiments}

Our experiments address three questions. First, does proprioceptive input
improve the prediction of the physical future, and do these improvements
translate into better plans (Sec.~\ref{sec:exp:physical})? Second, does Feel-WM
outperform visual-only navigation world models in closed-loop rough-terrain
navigation across wheeled and legged platforms (Sec.~\ref{sec:exp:closedloop})?
Third, can Feel-WM operate on a real robot with onboard planning in the field
(Sec.~\ref{sec:exp:real})?

\subsection{Experimental Setup}
\label{sec:exp:setup}

\paragraph{Robots and data}
Our closed-loop experiments use a wheeled Clearpath Husky, except for the legged
experiments, which use an ANYmal-C. For offline evaluation, we use
TartanDrive~1~\cite{triest2022tartandrive} and
TartanDrive~2~\cite{sivaprakasam2024tartandrive2}, both of which contain
off-road driving data collected with an all-terrain vehicle. We use these
datasets because they record the wheel and inertial signals our targets require
alongside the camera. Closed-loop simulation experiments use Gazebo and Isaac Sim.
We train separate Feel-WM models for each domain using the same architecture
and training procedure: one for each TartanDrive dataset, one for Gazebo and one
using the legged robot's walking data. For real-world deployment, we fine-tune
the Gazebo model on the real robot's own driving.

\paragraph{Evaluation metrics}
We report $R^2$ for proprioceptive feature prediction and AUC for failure-risk
prediction using the recorded hazard labels. For offline planning, we report
average displacement error (ADE) and relative trajectory error (RTE). For
closed-loop navigation, we report success rate (SR), route progress and two
ride-quality metrics: realized roughness, measured as the standard deviation of
vertical acceleration, and yaw vibration, measured as the mean absolute heading
rate computed from logged poses. Both show whether a model's success came with
more vibration or more heading change.

\begin{figure*}[t]
  \centering
  \includegraphics[width=0.85\linewidth]{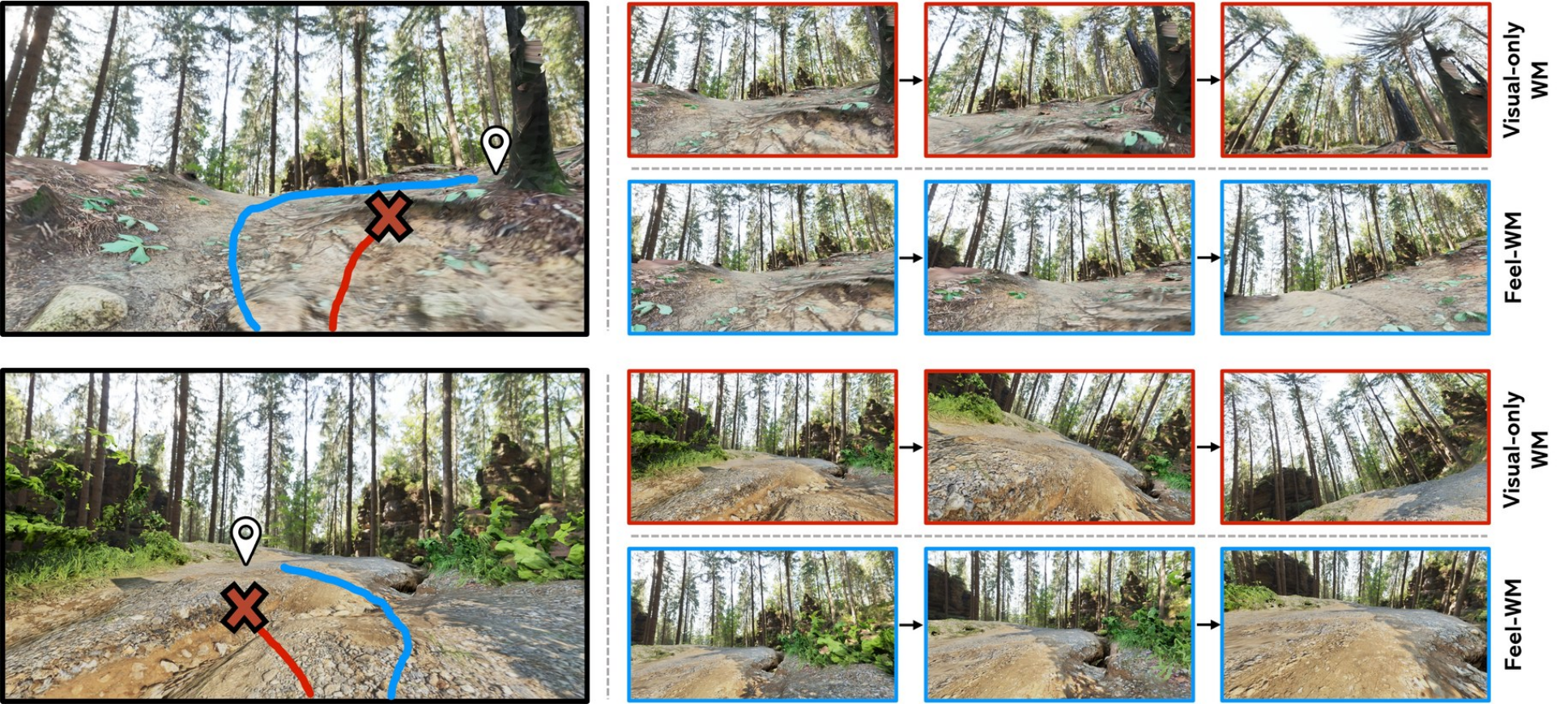}
    \caption{\textbf{Qualitative comparison in Isaac Sim.} Left: trajectories color-coded as
  visual-only model (red) and Feel-WM (blue); the pin denotes the goal, and $\times$ marks
  the rollover (top) and the body disturbance on exit (bottom). Right: frames along each
  episode.}
  \label{fig:isaac}
\end{figure*}

\subsection{Offline Evaluation on TartanDrive}
\label{sec:exp:physical}

\paragraph{Physical-future prediction}
To isolate the effect of the proprioceptive input, we train Feel-WM and
Feel-WM w/o proprio, which keeps the heads and drops the input, on
TartanDrive~2 and evaluate both on its validation split with three seeds
(Fig.~\ref{fig:physical}). Proprioceptive input raises the $R^2$ of the
predicted features and the AUC of the failure risk at every horizon shown, and
the gain holds on every seed.

\paragraph{Open-loop planning}
We compare against released checkpoints of three baselines:
NWM~\cite{bar2025navigation} and RAE-NWM~\cite{zhang2026rae}, which predict
future visual observations, and NoMaD~\cite{sridhar2024nomad}, a diffusion
policy that directly generates actions. For open-loop planning we use the
TartanDrive~1 model of Feel-WM on both datasets, so that Feel-WM, NWM and NoMaD
were all trained on TartanDrive~1, RAE-NWM on other navigation datasets, and no
model was trained on TartanDrive~2.
Following NWM's protocol, world-model planning selects candidate action
sequences based on the similarity between predicted future observations and the
goal image. All world models use the same CEM configuration and action prior
over a 3.2\,s horizon. We evaluate the resulting plans against recorded driver
trajectories (Table~\ref{tab:openloop}).

Among the world models, Feel-WM achieves the lowest ADE on both datasets.
NoMaD achieves a lower ADE on TartanDrive~2. This metric measures agreement
with the recorded driver trajectory and therefore aligns closely with NoMaD's
imitation objective. TartanDrive~1 records no failure labels, so its model
omits the failure-risk head and loss while keeping the proprioceptive input and
head; all world models here therefore plan on goal similarity alone. No plan
here is executed, so this evaluation does not assess the navigation benefit of
the failure-risk term.

\subsection{Closed-Loop Navigation in Simulation}
\label{sec:exp:closedloop}
We evaluate in two simulators. In Gazebo, we use the natural outdoor
environments of NEGS-UGV~\cite{sanchez2022automatically}, with three routes
spanning 110--165\,m through a flat park, an uneven forest and a steep, uneven
hill. In Isaac Sim, we use the outdoor environments of
TAO~\cite{yoo2024traversability}, reconstructed from real campus paths and
mountain trails: 13 environments with routes of 8--30\,m, divided by route
slope into 7 smooth and 6 rough. All models are trained on 880 Gazebo
trajectories totaling 14.0\,h, with no training on Isaac Sim. Evaluation in Isaac Sim
therefore tests zero-shot transfer.

\paragraph{Compared models}
We compare three models with a shared backbone architecture, training procedure, and training duration. The visual-only
world model conditions on camera observations and predicts future visual
features. Feel-WM additionally conditions on recent proprioceptive measurements
and predicts future proprioceptive features and failure risk. Feel-WM w/o
proprio keeps these heads and drops the proprioceptive input.

NWM and RAE-NWM were not trained in either simulator, and their diffusion
planners cost minutes per plan, or about a day of GPU time per episode at our
replanning rate. We therefore use a visual-only world model trained on the same
corpus as the controlled baseline for that paradigm.

\paragraph{Driving protocol}
Each model runs 50 episodes per route, following goal images recorded during a
single human-driven traversal. An episode is successful if
the robot reaches within 1\,m of the final goal. The planner evaluates 64
candidates per CEM iteration over two iterations and replans every four control
steps. Its eight-step rollout spans $3.4\,\mathrm{s}$ with step durations of
0.1, 0.5 and $1.0\,\mathrm{s}$. The visual-only model plans on goal similarity
alone, since it has no failure-risk head. Feel-WM and Feel-WM w/o proprio use
Eq.~\eqref{eq:score} with $\lambda_{\mathrm{fail}} = 1$.

\paragraph{Gazebo routes}
On the flat park route, all three models reach the goal in nearly every episode
(Table~\ref{tab:main}): flat ground gives the physical future nothing to add.
Performance differences emerge on more challenging terrain. On the uneven
forest route, Feel-WM achieves an SR of 84.0\,\%, compared with 54.0\,\% for the
visual-only model and 74.0\,\% for Feel-WM w/o proprio. On the steep, uneven
hill route, Feel-WM succeeds in 84.0\,\% against 58.0 and 68.0\,\%. The
comparison between Feel-WM and Feel-WM w/o proprio isolates the contribution of
proprioceptive conditioning. Feel-WM also achieves the lowest realized roughness
on all three routes and the lowest yaw vibration on the park and hill routes,
indicating that its higher success rate does not come at the expense of ride
quality.

Fig.~\ref{fig:gazebo} presents paired episodes for each route. The trajectories
nearly coincide in the park. In the forest and on the hill, the visual-only
model maintains its course through rough terrain and rolls over, whereas Feel-WM
detours around it.

\paragraph{Zero-shot in Isaac Sim}
In Isaac Sim, Feel-WM achieves the highest SR on rough terrain at 78.3\,\%, compared
with 67.7\,\% for the visual-only model and 62.0\,\% for Feel-WM w/o proprio.
On smooth terrain all three models succeed in most episodes, with Feel-WM
highest.

Fig.~\ref{fig:isaac} shows paired episodes from two environments. In the first,
a large tree root obstructs the direct path to the goal. The visual-only model
attempts to cross it and rolls over. Feel-WM predicts high failure risk for this
crossing and steers around the root to reach the goal. In the second, an eroded
depression runs along the center of the path. Both models succeed, but Feel-WM
traverses the shallow edge while the visual-only model crosses the deepest
section and undergoes a large body disturbance on exit.

The roughness metric shows little separation between models in Isaac Sim, with the
visual-only model achieving a slightly lower value. Feel-WM travels faster over
the same terrain, which raises its measured roughness. Feel-WM nevertheless
achieves the lowest yaw vibration in both terrain groups.

\begin{table}[t]
\centering
\small
\setlength{\tabcolsep}{5pt}
\resizebox{\ifdim\width>\linewidth\linewidth\else\width\fi}{!}{%
\begin{tabular}{@{}llccc@{}}
\toprule
Route & Method & SR & Prog. & Rollover \\
\midrule
\multirow{2}{*}{Rough}          & Visual-only WM & 31.0 & 43.7 & 51.0 \\
                                & Feel-WM        & 44.0 & 46.2 & 37.0 \\
\midrule
\multirow{2}{*}{Smooth}         & Visual-only WM & 52.5 & 75.3 & 33.3 \\
                                & Feel-WM        & 78.3 & 77.7 & 12.5 \\
\bottomrule
\end{tabular}}
\caption{\textbf{Closed-loop navigation on a legged robot.} We evaluate an ANYmal-C in
Isaac Sim over 6 smooth and 5 rough environments, 20 episodes each. All columns in \%.}
\label{tab:legged_cl}
\end{table}

\paragraph{Score ablation}
Planning the same Feel-WM checkpoint on the hill route with goal similarity
alone, Eq.~\eqref{eq:score} at $\lambda_{\mathrm{fail}} = 0$, lowers SR from
84.0 to 62.0\,\% and progress from 88.6 to 77.9\,\% over 50 episodes.
Predicted failure risk therefore improves trajectory selection on rough terrain.

\paragraph{Legged navigation}
We evaluate on a legged platform to assess whether the benefits of Feel-WM extend beyond
wheeled robots. We train the visual-only world model and Feel-WM on 815 trajectories
collected by an ANYmal-C in Isaac Sim. We replace the wheel slip of
Eq.~\eqref{eq:channels} with foot slip and add a cost-of-transport channel.
The robot walks with the pretrained rough-terrain
locomotion policy of Isaac Lab~\cite{mittal2023orbit}, which tracks the body-frame
velocity commands that the planner issues; both world models command the same policy.
Two of the 13 held-out environments are excluded as unsuitable for legged
locomotion, leaving six smooth and five rough. We evaluate each model over 20 episodes per environment and report rollover
rate in place of roughness (Table~\ref{tab:legged_cl}). Feel-WM achieves higher success
rates in both groups: 44.0\,\% versus 31.0\,\% on rough terrain and 78.3\,\% versus
52.5\,\% on smooth terrain. It also reduces rollover rates in both groups while
maintaining similar progress, indicating that its higher success rate is accompanied by
fewer falls.

\begin{figure}[t]
  \centering
  \includegraphics[width=\linewidth]{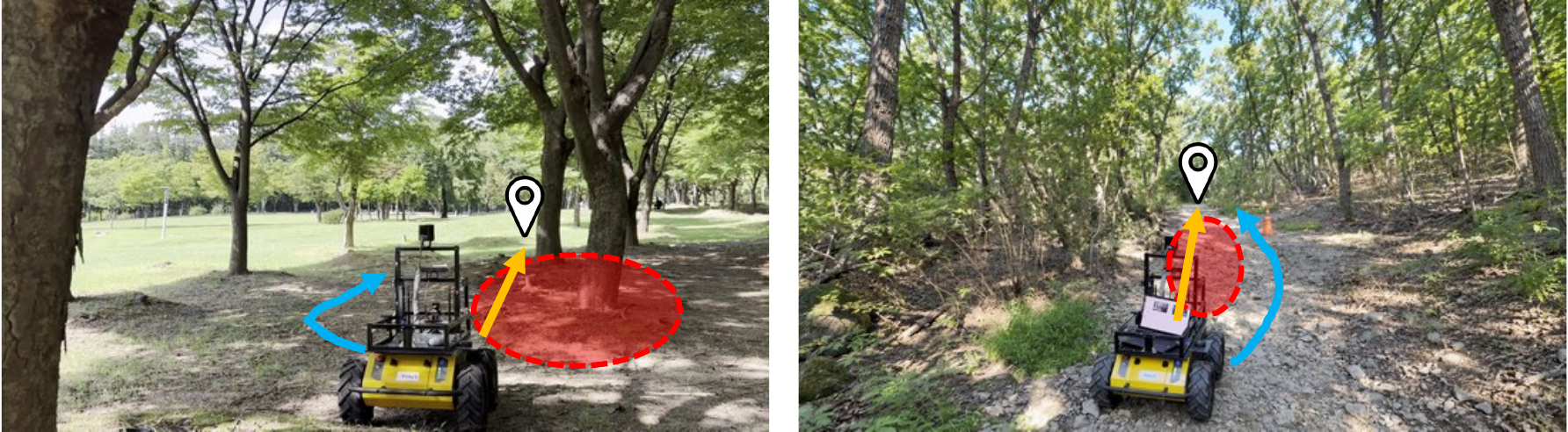}
    \caption{\textbf{Qualitative comparison on the real robot.} The pin denotes the next goal
  and the red region the hazard on the direct line. Arrows are color-coded as the
  trajectory driven by NoMaD (orange) and by Feel-WM (blue).}
  \label{fig:real}
\end{figure}

\begin{table}[t]
\centering
\small
\setlength{\tabcolsep}{5pt}
\resizebox{\ifdim\width>\linewidth\linewidth\else\width\fi}{!}{%
\begin{tabular}{@{}lcccccc@{}}
\toprule
\multirow{2}{*}{Method} & \multicolumn{2}{c}{Roots} & \multicolumn{2}{c}{Downhill} & \multicolumn{2}{c}{Uphill} \\
\cmidrule(lr){2-3}\cmidrule(lr){4-5}\cmidrule(lr){6-7}
 & SR & Rough.
 & SR & Rough.
 & SR & Rough. \\
\midrule
NoMaD & 60 & 0.303
      & 30 & 0.731
      & 50 & 0.783 \\
Feel-WM  & 90 & 0.261
      & 70 & 0.593
      & 80 & 0.612 \\
\bottomrule
\end{tabular}}
\caption{\textbf{Closed-loop navigation on the real robot.} We evaluate a Clearpath
Husky on three courses, 10 runs per course and method. SR in \%, roughness in m/s$^2$.}
\label{tab:realworld}
\end{table}

\subsection{Real-World Deployment}
\label{sec:exp:real}

\paragraph{Robot platform}
We deploy Feel-WM on a Clearpath Husky with a ZED~2i stereo camera and an IMU, planning
onboard a Jetson AGX Orin. We fine-tune Feel-WM from its Gazebo model and NoMaD from its released
checkpoint. Both use under an hour of the Husky's own driving, recorded on
trails away from the evaluation courses.

\paragraph{Onboard planning}
We fine-tune Feel-WM with a perceiver
resampler~\cite{alayrac2022flamingo} that compresses the 256 patch tokens of each frame to
32 latents before the dynamics transformer and restores the grid after it, so the heads and
the rollout keep full resolution while the transformer runs on an eighth of the tokens.
Onboard the planner evaluates 16 candidates over two CEM iterations and replans every
0.8\,s. The world-model baselines are not deployed: NWM reports 30\,s to simulate a
single trajectory on a desktop GPU~\cite{bar2025navigation}, and RAE-NWM rolls out with
50 ODE steps per frame~\cite{zhang2026rae}, far outside an onboard replanning budget.

\paragraph{Field results}
We drive three courses, 10 runs per course and method
(Table~\ref{tab:realworld}). NoMaD follows a goal-image graph built from the same pass. A
run counts as a success only if it reaches the goal with no operator intervention. Feel-WM completes 24 of the 30 runs against 14 for NoMaD and rides smoother on every
course.
Fig.~\ref{fig:real} shows two of these runs. A rock field lies on the direct line to the
next goal; NoMaD holds that line and drives across it, while Feel-WM passes beside it and
rejoins the taught route beyond.
\section{Conclusion}
\label{sec:conclusion}

We presented Feel-WM, a navigation world model that conditions on the robot's
proprioceptive state to predict the physical future alongside the scene, together
with a planner that weighs the predicted failure risk against goal similarity.
Feel-WM outperforms visual-only navigation world models in open-loop planning on
recorded off-road driving data and in closed-loop rough-terrain navigation across
wheeled and legged robots. Deployed on a Clearpath Husky, it completes off-road
courses that an end-to-end policy fails.

Two limitations motivate future work. First, Feel-WM learns the physical future
solely from recorded driving data, whose limited volume and diversity constrain
generalization to unseen terrain. World models that learn from experience share
this dependence on training coverage. One direction is to train a visual encoder
on off-road driving data with objectives that capture terrain properties relevant
to the robot--terrain interaction. Second, onboard planning constrains model size
and rollout computation, creating a trade-off between prediction accuracy and
computational cost. Future work on this trade-off includes generative rollouts
efficient enough for onboard sampling, more compact frame representations, and
models that jointly predict actions and future scenes.


\bibliographystyle{IEEEtran}
\bibliography{ref}

@INPROCEEDINGS{RothP-RSS-25,
    AUTHOR    = {Pascal Roth AND Jonas Frey AND Cesar Cadena AND Marco Hutter},
    TITLE     = {Learned perceptive forward dynamics model for safe and platform-aware robotic navigation},
    BOOKTITLE = {Proc. Robot.: Sci. Syst. (RSS)},
    YEAR      = {2025}
}

@article{zhao2026learning,
  title={Learning When to Jump for Off-road Navigation},
  author={Zhao, Zhipeng and Fu, Taimeng and Su, Shaoshu and Du, Qiwei and Esfahani, Ehsan Tarkesh and Dantu, Karthik and Chowdhury, Souma and Wang, Chen},
  journal={arXiv:2602.00877},
  year={2026}
}

@inproceedings{bar2025navigation,
  title={Navigation world models},
  author={Bar, Amir and Zhou, Gaoyue and Tran, Danny and Darrell, Trevor and LeCun, Yann},
  booktitle={Proc. IEEE/CVF Conf. Comput. Vis. Pattern Recognit. (CVPR)},
  pages={15791--15801},
  year={2025}
}

@inproceedings{nie2025wmnav,
  title={{WMNav}: Integrating vision-language models into world models for object goal navigation},
  author={Nie, Dujun and Guo, Xianda and Duan, Yiqun and Zhang, Ruijun and Chen, Long},
  booktitle={Proc. IEEE/RSJ Int. Conf. Intell. Robots Syst. (IROS)},
  pages={2392--2399},
  year={2025}
}

@article{hu2026navthinker,
  title   = {{NavThinker}: Action-Conditioned World Models for Coupled Prediction and Planning in Social Navigation},
  author  = {Hu, Tianshuai and Gong, Zeying and Kong, Lingdong and Mei, XiaoDong and Ding, Yiyi and Zeng, Qi and Liang, Ao and Li, Rong and Zhong, Yangyi and Liang, Junwei},
  journal = {arXiv:2603.15359},
  year    = {2026}
}

@inproceedings{castro2023does,
  title={How does it feel? {S}elf-supervised costmap learning for off-road vehicle traversability},
  author={Castro, Mateo Guaman and Triest, Samuel and Wang, Wenshan and Gregory, Jason M and Sanchez, Felix and Rogers, John G and Scherer, Sebastian},
  booktitle={Proc. IEEE Int. Conf. Robot. Autom. (ICRA)},
  pages={931--938},
  year={2023}
}

@INPROCEEDINGS{frey23fast,
  AUTHOR    = {Jonas Frey AND Matias Mattamala AND Nived Chebrolu AND Cesar Cadena AND Maurice Fallon AND Marco Hutter},
  TITLE     = {Fast traversability estimation for wild visual navigation},
  BOOKTITLE = {Proc. Robot.: Sci. Syst. (RSS)},
  YEAR      = {2023}
}

@article{elnoor2024pronav,
  title={{ProNav}: Proprioceptive traversability estimation for legged robot navigation in outdoor environments},
  author={Elnoor, Mohamed and Sathyamoorthy, Adarsh Jagan and Weerakoon, Kasun and Manocha, Dinesh},
  journal={IEEE Robot. Autom. Lett.},
  volume={9},
  number={8},
  pages={7190--7197},
  year={2024}
}

@InProceedings{shah2023vint,
  title = 	 {{ViNT}: A Foundation Model for Visual Navigation},
  author =       {Shah, Dhruv and Sridhar, Ajay and Dashora, Nitish and Stachowicz, Kyle and Black, Kevin and Hirose, Noriaki and Levine, Sergey},
  booktitle = 	 {Proc. Conf. Robot Learn. (CoRL)},
  pages = 	 {711--733},
  year = 	 {2023}
}

@inproceedings{sridhar2024nomad,
  title={{NoMaD}: Goal masked diffusion policies for navigation and exploration},
  author={Sridhar, Ajay and Shah, Dhruv and Glossop, Catherine and Levine, Sergey},
  booktitle={Proc. IEEE Int. Conf. Robot. Autom. (ICRA)},
  pages={63--70},
  year={2024}
}

@InProceedings{zhou2025dino-wm,
  title = 	 {{DINO}-{WM}: World Models on Pre-trained Visual Features enable Zero-shot Planning},
  author =       {Zhou, Gaoyue and Pan, Hengkai and LeCun, Yann and Pinto, Lerrel},
  booktitle = 	 {Proc. Int. Conf. Mach. Learn. (ICML)},
  pages = 	 {79115--79135},
  year = 	 {2025}
}

@article{zhang2025efficient,
  title={Efficient image-goal navigation with representative latent world model},
  author={Zhang, Zhiwei and Zhang, Hui and Huang, Kaihong and Shi, Chenghao and Lu, Huimin},
  journal={arXiv:2511.11011},
  year={2025}
}

@article{zhang2026rae,
  title={{RAE-NWM}: Navigation World Model in Dense Visual Representation Space},
  author={Zhang, Mingkun and Shen, Wangtian and Zhang, Fan and Qin, Haijian and Pei, Zihao and Meng, Ziyang},
  journal={arXiv:2603.09241},
  year={2026}
}

@article{yoo2024traversability,
  title={Traversability-aware adaptive optimization for path planning and control in mountainous terrain},
  author={Yoo, Se-Wook and Son, E-In and Seo, Seung-Woo},
  journal={IEEE Robot. Autom. Lett.},
  volume={9},
  number={6},
  pages={5078--5085},
  year={2024}
}

@article{azuma2026navwam,
  title={{NavWAM}: A Navigation World Action Model for Goal-Conditioned Visual Navigation},
  author={Azuma, Daichi and Miyanishi, Taiki and Sakamoto, Koya and Kurita, Shuhei and Zhu, Yaonan and Khrapchenkov, Petr and Kawanabe, Motoaki and Iwasawa, Yusuke and Matsuo, Yutaka},
  journal={arXiv:2606.13494},
  year={2026}
}

@article{yang2026wam,
  title={{WAM-Nav}: Asymmetric latent world-action modeling for unified visual navigation},
  author={Yang, Ning and Huang, Yan and Peng, Kaiwen and He, Ziheng and Wang, Kai and Miao, Cui and Lyu, Kailin and Li, Guo and Wang, Xiaofeng and Zhu, Zheng and others},
  journal={arXiv:2606.04907},
  year={2026}
}

@article{yang2026terrainformer,
  title={{TerrainFormer}: World Model-Guided Decision Transformer for Autonomous Off-Road Navigation},
  author={Yang, Yongzhi and Ricks, Kenneth},
  journal={Sensors},
  volume={26},
  number={12},
  pages={3795},
  year={2026}
}

@article{sanchez2022automatically,
  title={Automatically annotated dataset of a ground mobile robot in natural environments via gazebo simulations},
  author={S{\'a}nchez, Manuel and Morales, Jes{\'u}s and Mart{\'\i}nez, Jorge L and Fern{\'a}ndez-Lozano, Juan Jes{\'u}s and Garc{\'\i}a-Cerezo, Alfonso},
  journal={Sensors},
  volume={22},
  number={15},
  pages={5599},
  year={2022}
}

@article{oquab2024dinov,
title={{DINO}v2: Learning Robust Visual Features without Supervision},
author={Maxime Oquab and Timoth{\'e}e Darcet and Th{\'e}o Moutakanni and Huy V. Vo and Marc Szafraniec and Vasil Khalidov and Pierre Fernandez and Daniel Haziza and Francisco Massa and Alaaeldin El-Nouby and Mido Assran and Nicolas Ballas and Wojciech Galuba and Russell Howes and Po-Yao Huang and Shang-Wen Li and Ishan Misra and Michael Rabbat and Vasu Sharma and Gabriel Synnaeve and Hu Xu and Herve Jegou and Julien Mairal and Patrick Labatut and Armand Joulin and Piotr Bojanowski},
journal={Trans. Mach. Learn. Res.},
year={2024}
}

@inproceedings{perez2018film,
  title={{FiLM}: Visual reasoning with a general conditioning layer},
  author={Perez, Ethan and Strub, Florian and De Vries, Harm and Dumoulin, Vincent and Courville, Aaron},
  booktitle={Proc. AAAI Conf. Artif. Intell.},
  volume={32},
  year={2018}
}

@article{lee2024learning,
  title={Learning vehicle dynamics from cropped image patches for robot navigation in unpaved outdoor terrains},
  author={Lee, Jeong Hyun and Choi, Jinhyeok and Ryu, Simo and Oh, Hyunsik and Choi, Suyoung and Hwangbo, Jemin},
  journal={IEEE Robot. Autom. Lett.},
  volume={9},
  number={5},
  pages={4035--4042},
  year={2024}
}

@inproceedings{triest2024velociraptor,
  title={{Velociraptor}: Leveraging visual foundation models for label-free, risk-aware off-road navigation},
  author={Triest, Samuel and Sivaprakasam, Matthew and Aich, Shubhra and Fan, David and Wang, Wenshan and Scherer, Sebastian},
  booktitle={Proc. Conf. Robot Learn. (CoRL)},
  year={2024}
}

@article{alayrac2022flamingo,
  title={Flamingo: a visual language model for few-shot learning},
  author={Alayrac, Jean-Baptiste and Donahue, Jeff and Luc, Pauline and Miech, Antoine and Barr, Iain and Hasson, Yana and Lenc, Karel and Mensch, Arthur and Millican, Katherine and Reynolds, Malcolm and others},
  journal={Adv. Neural Inf. Process. Syst.},
  volume={35},
  pages={23716--23736},
  year={2022}
}

@article{chen2025astranav,
  title={{AstraNav-World}: World model for foresight control and consistency},
  author={Chen, Jintao and Hu, Junjun and Bai, Haochen and Luo, Minghua and Xue, Xinda and Ren, Botao and Bai, Chengyu and Xie, Shichao and Chen, Ziyi and Liu, Fei and others},
  journal={arXiv:2512.21714},
  year={2025}
}

@article{shen2026efficient,
  title={An efficient and multi-modal navigation system with one-step world model},
  author={Shen, Wangtian and Meng, Ziyang and Ma, Jinming and Zhou, Mingliang and Xiang, Diyun},
  journal={arXiv:2601.12277},
  year={2026}
}

@article{kim2026canonnav,
  title={{CanonNav}: Disentangling Navigation Behavior from Camera Geometry in Cross-Platform Visual Navigation},
  author={Kim, Dong-Wook and Hwang, Ji-Hoon and Son, E-In and Oh, Mintaek and Seo, Seung-Woo},
  journal={arXiv:2608.30242},
  year={2026}
}

@inproceedings{triest2022tartandrive,
  title={{TartanDrive}: A large-scale dataset for learning off-road dynamics models},
  author={Triest, Samuel and Sivaprakasam, Matthew and Wang, Sean J and Wang, Wenshan and Johnson, Aaron M and Scherer, Sebastian},
  booktitle={Proc. IEEE Int. Conf. Robot. Autom. (ICRA)},
  pages={2546--2552},
  year={2022}
}

@inproceedings{sivaprakasam2024tartandrive2,
  title={{TartanDrive} 2.0: More modalities and better infrastructure to further self-supervised learning research in off-road driving tasks},
  author={Sivaprakasam, Matthew and Maheshwari, Parv and Castro, Mateo Guaman and Triest, Samuel and Nye, Micah and Willits, Steve and Saba, Andrew and Wang, Wenshan and Scherer, Sebastian},
  booktitle={Proc. IEEE Int. Conf. Robot. Autom. (ICRA)},
  doi={10.1109/ICRA57147.2024.10611265},
  year={2024}
}

@article{deboer2005tutorial,
  title={A tutorial on the cross-entropy method},
  author={de Boer, Pieter-Tjerk and Kroese, Dirk P. and Mannor, Shie and Rubinstein, Reuven Y.},
  journal={Ann. Oper. Res.},
  volume={134},
  number={1},
  pages={19--67},
  year={2005}
}

@article{mittal2023orbit,
  title={Orbit: A unified simulation framework for interactive robot learning environments},
  author={Mittal, Mayank and Yu, Calvin and Yu, Qinxi and Liu, Jingzhou and Rudin, Nikita and Hoeller, David and Yuan, Jia Lin and Singh, Ritvik and Guo, Yunrong and Mazhar, Hammad and Mandlekar, Ajay and Babich, Buck and State, Gavriel and Hutter, Marco and Garg, Animesh},
  journal={IEEE Robot. Autom. Lett.},
  volume={8},
  number={6},
  pages={3740--3747},
  year={2023}
}

\end{document}